# Cross-Domain Bilingual Lexicon Induction via Pretrained Language Models

**Qiuyu Ding, Zhiqiang Cao***, **Hailong Cao, Tiejun Zhao**
School of Computer Science of Technology, Harbin Institute of Technology
qiuyuding@stu.hit.edu.cn, Zhiqiang_Cao@stu.hit.edu.cn,
caohailong@hit.edu.cn, tjzhao@hit.edu.cn

## Abstract

Bilingual Lexicon Induction (BLI) is generally based on common domain data to obtain monolingual word embedding, and by aligning the monolingual word embeddings to obtain the cross-lingual embeddings which are used to get the word translation pairs. In this paper, we propose a new task of BLI, which is to use the monolingual corpus of the general domain and target domain to extract domain-specific bilingual dictionaries. Motivated by the ability of Pre-trained models, we propose a method to get better word embeddings that build on the recent work on BLI. This way, we introduce the Code Switch(Qin et al., 2020) firstly in the cross-domain BLI task, which can match different strategies in different contexts, making the model more suitable for this task. Experimental results show that our method can improve performances over robust BLI baselines on three specific domains by averagely improving 0.78 points.

## 1 Introduction

Bilingual Lexicon Induction (BLI) aims to find a bilingual dictionary based on word embeddings of two languagesArtetxe et al. (2018); Lample et al. (2018); Artetxe et al. (2016). Generally, most BLI methods are researched in a general domain(Artetxe et al., 2019; Cao et al., 2019; Zhang et al., 2021; Artetxe et al., 2018; Patra et al., 2019), but the specific domain is neglected. However, the translation of words for professional fields has application value that cannot be ignored, because these words are often professional, and are frequently be used in the field these belonged to. Therefore, research in this area is more practical.

At present, most BLI methods carry out word mapping based on the space of word embeddings, and word embeddings are trained from large-scale monolingual corpus(Ruder et al., 2019). However, it is yet to be seen whether these methods are suitable for bilingual lexicon extraction in professional fields. As we can see in table 1, the classic and efficient BLI approach, Muse and Vecmap, perform much worse on the Medical dataset than on the Wiki dataset. On one hand, the specialized domain data set is relatively smaller compared to the generic domain data set generally, and specialized words have a lower frequency, which will directly affect the translation quality of bilingual dictionaries. On the other hand, static word embeddings are widely used for BLI, however, in some specific fields, the meaning of words is greatly influenced by context, in this case, using only static word embeddings may lead to greater bias.

*Corresponding Author.

|  | Wiki | | Medicine | |
| --- | --- | --- | --- | --- |
|  | en-zh | zh-en | en-zh | zh-en |
| Muse | 31.06 | 33.6 | 0 | 17.95 |
| Vecmap | 32.43 | 41.85 | 27.54 | 18.82 |

Table 1: Word translation accuracy of Muse and Vecmap in general domain and medical domain

To this end, we propose a novel BLI method, which is based on combing static word embeddings and contextual word representations, and introduce the Code Switch method enables domain terms to make better use of contextual information across languages. In particular, there are different Code Switch strategies in different sentences, to ensure that context information can be used more appropriately. In summary, our main contributions are as follows:

- A novel BLI task is proposed to extract cross-domain bilingual dictionaries, which can use the monolingual corpus of general and target domains.

- Utilize the ability of Pre-trained models naturally, and propose to introduce the Code Switch() method firstly into BLI task, and combine context-based word replacement strategies in this model to make it more suitable for the BLI task, and improve the effectiveness of the model training on domain data to get better word embeddings.

## 2 Background

We can simply divide the existing BLI methods based on static word embeddings(Artetxe et al., 2018; Lample et al., 2018) and based on contextual contextual word representations(Aldarmaki and Diab, 2019; Wang et al., 2019; Cao et al., 2019) according to the type of word embeddings used, until CSCBLI Zhang et al. (2021) puts forward another idea, that is, combining the contextual word representations(Conneau and Lample, 2019; Devlin et al., 2018; Conneau et al., 2019) and static word embeddings(Xing et al., 2015; Artetxe et al., 2016; Smith et al., 2017) when dealing with BLI tasks. It consists of the following components.

**The static word embeddings acquisition.**

CSCBLI gets the static word embeddings from Vecmap(Artetxe et al., 2018), which is one of the representatives and robust static word embedding-based methods. First of all, the monolingual word embeddings are obtained by training separately from a large-scale monolingual corpus(Ruder et al., 2019). Let X is word embeddings of the source language and Y is words embeddings of the target language, Vecmap aims to learn the linear transformation matrices $W_X$ and $W_Y$, which can let the mapped embeddings $XW_X$ and $YW_Y$ in same cross-lingual space, specifically, it includes two important parts, unsupervised initialization, and self-learning.

**Unsupervised initialization.** On the premise that the two language space matrices are isometry, first sort $M_X$ and $M_Z$ according to the value of each row, and then for a word X, can search the corresponding $M_Z$ according to the row where the $M_X$ after sorted is located to get the translation result.

**Self-learing.** First calculate the orthogonal mapping of the initial dictionary such that the optimal solutions $W_X = Z$ and $W_Z = V$ maximize the similarity, and then calculate the optimal dictionary on the similarity matrix of the mapping embedding by retrieving the nearest neighbors from the source language to the target language.

**The contextual representation acquisition.**

Contextual representations are generally obtained from Pre-trained models(Kenton and Toutanova; Conneau and Lample, 2019), and a word could have different word representations in different contexts. Since a word generally appears more than once in the training corpus, there are methods proposed with an average anchor method to solve the problem that the same word has different word representations without using a parallel corpus. Let the contextual word representation of a word $x$ in context $c_i$ be denoted as $r_{x,c_i}$, when x appears p times in the corpus, the average anchor of x in these contexts can be represented as:

$$a_x = \frac{\sum_{i=1}^{p} r_{x,c_i}}{p} \quad (1)$$

**The unified word representation acquisition.**

Due to the strong assumption that the source language space and the target language space are orthogonal in the process of obtaining static word embeddings, this assumption may not be necessary for all conditions(Zhang et al., 2017; Søgaard et al., 2018), which will lead to some words being still far away from the correct translation result after mapping. In order to pull words closer to their translations in the target language, CSCBLI proposed a spring network that can pull the mapped word embeddings to better positions, and the training of this spring network used adversarial training.

Regarding the adversarial training of the spring network, CSCBLI proposes two methods, supervised and unsupervised. Since the setting of this article is based on the domain monolingual corpus for BLI, only unsupervised adversarial training is introduced here.

First, use the unsupervised BLI methods to obtain $I$ parallel word translation pairs, let $u_x^i$ and $u_y^i$ are the unified representations corresponding to the $i_t h$ entry of the bilingual dictionary given by Vecmap, and the loss of unsupervised adversarial training is:

$$L_{sup} = -\sum_{i=1}^{\geq I} J \times \cos(u_x^i, u_y^i) - \sum_{j=1}^{J} \cos(u_x^i, w_{y^-}^j) \quad (2)$$

In this process, $(u_x^i, u_y^i)$ is the positive translation pair, and $(u_x^i, w_{y^-}^j)$ is the negative pair which $y^-$ is not the correct translation of $x$, and for a word $x$ select $J$ negative pairs. During training, the Cosine similarity of positive translation pairs will gradually increase, and the Cosine similarity of negative translation pairs will gradually increase and will gradually become smaller.

**Similarity interpolation.**

The similarity interpolation between static and contextual embeddings is used to creating new word embeddings. Since both the unified word representation space and the mapped contextual word embedding space can calculate the similarity between words, we have the following formula:

$$S = \cos(u_x, u_y) + \lambda \cos(a'_x, a'_y) \qquad (3)$$

Where $x$ is a word in the source language, $y$ is a word in the target language, $u_x$ and $u_y$ are the unified word representations of $x$ and $y$, $a'_x$ and $a'_y$ are the mapped contextual representations, calculate the similarity between them respectively, $\lambda$ is the weight coefficient, which is tuned by an unsupervised procedure: when source-to-target model and target-to-source model have been trained, the word $x$ in the validation set is aligned to $y'$ based on equation (3), then $y'$ is back aligned to $x$ based on the inverse version of equation (3). By calculating the $S$ of each word in the target language of $x$, the largest one is finally selected as the translation result of $x$.

## 3 Methods

The work of CSCBL proves that the combination of contextual word representation and static word embedding can improve the accuracy of bilingual dictionary extraction. However, there are still parts that can be improved at the stage of obtaining contextual word representations, making it more suitable for cross-domain BLI tasks. Therefore, We added the Code Switch method in the fine-tuning stage of the Pre-trained model based on the domain corpus, supplemented by the context-based word replacement strategies, so that the context information of the target language is naturally included when the model is pre-trained.

### 3.1 Model training

The acquisition of contextual word embeddings mainly includes three steps, respectively model training, acquisition of contextual word representation, and contextual word representation alignment.

The first step of dynamic word embeddings obtain is the Pre-trained model fine-tuning on the domain datasets. In this training process, we joined the Code Switch method to improve the Pre-trained model's training process and let it more suitable for the BLI task.

**Code Switch.** To put it simply, the Code Switch method is to randomly replace words with the help of an external landing dictionary during the Pre-trained model's training step or fine-tuning step, and replace the word of the source language with the corresponding translation of the target language.

Code Switch includes three steps mainly. The first step is randomly select sentences for word replacement, that is in a given batch of training data, we randomly select some sentences for the next operation. This step can ensure that in a batch some sentences are modified, while other parts of the sentences remain the same without modification. Secondly, in these selected sentences randomly select some words. Thirdly, translate selected words in selected sentences into the target language, that is find the corresponding word translation from the external bilingual dictionary to replace these words.

**Word Replacement Strategy.** In order to make the Code Switch method more suitable for cross-domain BLI tasks, we add some replacement strategies based on it. We can simply divide vocabularies in a sentence into two groups, general vocabulary, and domain vocabulary. For the classification of these two groups, we can adopt a simple strategy, that is, if the relative word frequency in the general corpus is lower than a certain frequency, we consider it to be a domain word, otherwise we consider it a general word. For example, let the frequency of word $x$ in a general domain training set is $F_{Gx}$, the total word number of this training set is $G$, so the relative word frequency of $x$ can be described as $F_{Gx}/G$, similarly, let the word frequency of $x$ in a specific domain training set is $F_{Dx}$, the total word numbers of this training set is $D$, so the frequency of $x$ in it can be described as $F_{Dx}/D$. When equation 4 holds, we consider x to be a general vocabulary, otherwise, it is a domain vocabulary.

$$F_{Gx}/G \geq F_{Dx}/D \qquad (4)$$

Based on Code Switch, after selecting some sentences in a batch, for the rest sentences, if the ratio of domain vocabulary reaches the threshold $\gamma$, all words in this sentence will be replaced, otherwise, it will still be carried out random word translation.

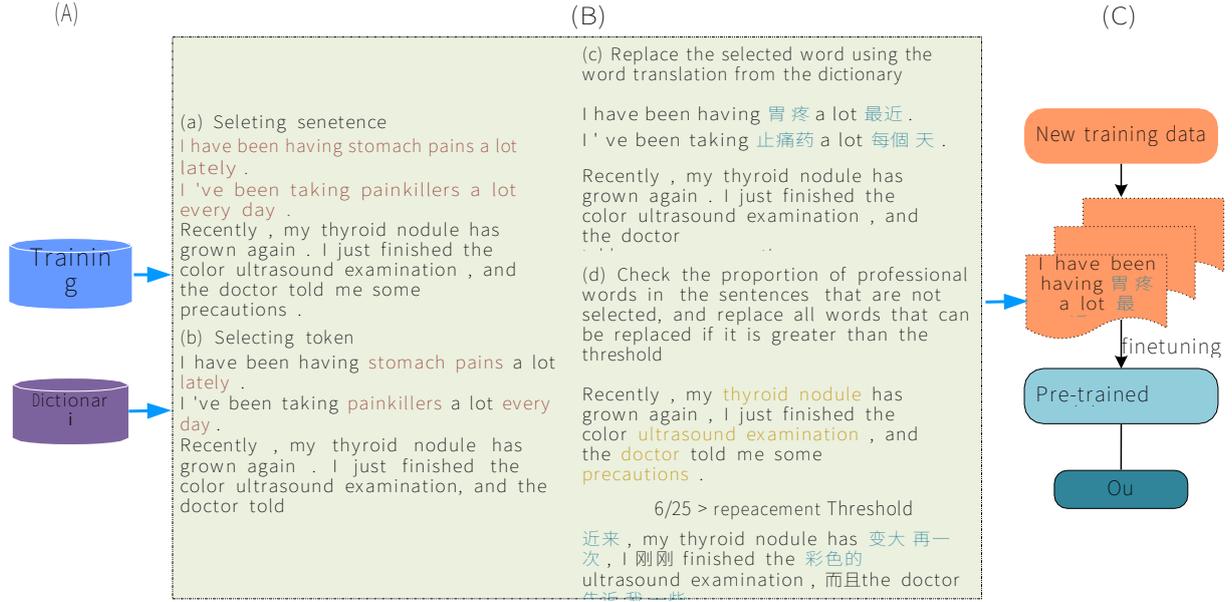

Figure 1: Illustration of Pre-trained model training process with the Code-switch combing words replacement strategies.

Algorithm 1 shows the pseudo-code of the new Code Switch which added word replacement strategies we proposed in this task.

**Algorithm 1:** Optimized Code Switch method framework.

**Input:** Source language training data: $S$, a set of billigual dictionaries: $d$, sentences replace ratio: $\alpha$, words replacement ratio: $\beta$, domain words replacement ratio: $\gamma$
**Output:** Code Switched training data: $S_{new}$
1 **for** $s$ *in* $S$ **do**  // $s$ is a sentence in $S$
2   **while** $random() < \alpha$ **do**
3     **while** $T_{word}$ [SEP] **do**
4       **if** $random() < \beta$ **then**
5         $S_{word} \leftarrow dict_{s\text{-}t}[T_{word}]$
6       **else**
7         $S_{word} \leftarrow S_{word}$
8     **end**
9   **while** $random() > \alpha$ **do**
10     **if** *relative p of domain words* $> \gamma$ **then**
11       $S_{word} \leftarrow dict_{s\text{-}t}[T_{word}]$
12

### 3.2 Word embeddings acquisition

**Contextual word representations extraction.** After model trainings, the contextual representations can be obtained through the Pre-trained model. Due to a word may appear in many sentences, we use an average anchor to be the representation of a word. That is, find the context in which the word is used first, get the lexical representation of that word in all the contexts, and calculate an average of these. Since some words appear in a large number of contexts, so we randomly select some of them for the average anchor calculation in this condition.

For each word in the vocabulary of the source language and target language to obtain the representation, we use the average anchor method to obtain its contextual word representation. Then, the classical and more robust Vecmap method is used to align the contextual word representations obtained in the two languages into one same space. In this way, the contextual word embeddings after aligning the contextual word representation obtained from the Pre-trained model are realized.

**Static word embeddings extraction.** The mapped static word embeddings are obtained from Vecmap, The general process is as follows

First, obtain its word embeddings based on monolingual corpora of two languages, such as using Word2Vec, FASTTEXT(Bojanowski et al., 2017). Then a linear transformation is learned based on the word embeddings of the two lan-guages such that $XW_x$ and $ZW_z$ are in the same cross-lingual space. The process of learning a linearly varying matrix is divided into four steps, which are normalizing the word embeddings, initialization which can create an initial solution, self-learning stage which can improve the solution iteratively, and refinement step which can further improve the resulting mapping.

In this way, by calculating the similarity between word embeddings, the translation results of the words in the source language in the target language can be obtained.

After that, a spring network is used to pull the static embeddings: let the similar words be closer, and different words are farther, which is trained by the contextual representations network. After that, we got the new static word embeddings.

## 4 Experimental settings

|  | wiki | | Medicine | |
| --- | --- | --- | --- | --- |
|  | En | Zh | En | Zh |
| sentences | 72494 | 313 | 3000 | 48821 |
| words | 2853268 | 15834 | 529661 | 657983 |
|  | Law | | Financial | |
|  | En | Zh | En | Zh |
| sentences | 681 | 15834 | 1776 | 548 |
| words | 20591 | 313 | 221862 | 25042 |

Table 2: Details of datesets. We shows the number(k) of words and sentences in all the monolingual corpora used. The word number is the total number of words contained in the preprocessed corpus.

We test our method in the unsupervised BLI task in English-to-Chinese (en-zh) and Chinese-to-English (zh-en) directions on three domain datasets.

### 4.1 Data

Monolingual corpora are needed to get the contextual representations and static word embeddings. In this setting, we only use monolingual corpus in the domain, however, since the domain monolingual corpus is relatively small, a general corpus is needed to give a good initialization. For this reason, we use the latest Wikipedia[1] to give better static monolingual embeddings, trained by the FASTTEXT. Besides, we use it to finetune the XLM model. In this process, the training data used for contextual word representations and static word embeddings are the same.

As for the monolingual corpus for domains, as shown in table 2, we collect public datasets from multiple parties, as well as download them from public websites and stitch them together. In total, we collected corpora in three domains, including medicine_en[2], medicine_zh(Liu et al., 2020), law_en(Galgani and Hoffmann, 2010), law_zh[3], and finance_en(Ding et al., 2014), finance_zh(Li and Sun, 2007).

As for the test data sets, since there are few public word translation test sets on domain datasets, we follow the method of the Muse test set construction, which is to construct a test set based on word frequency. Since there are few public word translation test sets on domain datasets, we follow the method of the Muse test set construction, which is to construct a test set based on word frequency. Since the domain dataset is much smaller than the general dataset (wiki), we appropriately reduced the size of parallel word pairs including train data and test data. The 5000 pairs of train data and 1500 pairs of test data used by the Muse change to 2000 pairs of train data and 800 pairs of test data. Regarding the word translation from the source language to the target language, we use the translation results given by various translation engines (including Google Translate, Youdao Translate, Baidu Translate, Bing Translate, and Sogou Translate) to construct word pairs by voting, which is relatively simple, and in the case of multiple translations of a word, it can be reserved for multiple translations with the same number of top-ranked translations.

### 4.2 Baseline Systems

We compare our method with four popular BLI systems: Muse(include supervised and unsupervised), Vecmap(include supervised and unsupervised), BLISS, and CSCBLI. All these systems are run with the default hyper-parameters settings.

- **Muse:** We conduct experiments on both supervised and unsupervised methods provided by the Muse and test the performance of the Muse using Cosine and CSLS two distance metrics. The supervised method of the Muse mainly uses the grounded seed dictionary for initialization and then uses Procrustes analysis to perform multiple iterations to achieve word alignment. The unsupervised method of the Muse is to use adversarial training for initialization, get some roughly aligned word pairs, and then use Procrustes analysis to iteratively align. Seed Dictionaries in Supervised Methods We use all ground truth dictionaries contained in the English and Chinese language pairs provided by the Muse.

---
[1] https://github.com/attardi/wikiextractor
[2] https://ufal.mff.cuni.cz/ufal_medical_corpus
[3] https://flk.npc.gov.cn/fl.html

|  | Medicine | | Law | | Financial | | avg |
|---|---|---|---|---|---|---|---|
|  | En-zh | Zh-en | En-zh | Zh-en | En-zh | Zh-en |  |
| Muse_unsupervised$_C$SLS | 0 | 17.95 | 0 | 18.92 | 0 | 19.74 | 9.44 |
| Muse_supervised$_C$SLS | 19.64 | 16.52 | 11.73 | 14.50 | 18.19 | 16.15 | 16.12 |
| Vecmap_unsupervised$_C$SLS | 27.54 | 18.82 | 13.33 | 14.63 | 30.54 | 23.59 | 21.41 |
| Vecmap_supervised$_C$SLS | 22.93 | 13.51 | 14.00 | 11.10 | 19.92 | 13.08 | 15.76 |
| CSCBLI | 29.85 | 23.48 | 28.68 | 21.85 | 32.87 | 25.78 | 27.08 |
| CSCBLI_codeSwitch | **30.68** | **24.12** | **29.65** | **22.78** | **33.65** | **26.18** | **27.84** |

Table 3: The accuracy of P@1 on medicine, law, and financial three domains on language pair of en, zh. Unsupervised and supervised refer to whether the method uses a seed dictionary, and the distance between words is measured using CSLS. Considering the randomness of the Muse method, we ran three seeds for each experiment of Muse, and then select the optimal value as the final result.

|  | Components Used | | Medicine | | Law | | Financial | | avg |
|---|---|---|---|---|---|---|---|---|---|
|  | Code_switch | word replacement | en-zh | zh-en | en-zh | zh-en | en-zh | zh-en |  |
| #0 | ✗ | ✗ | 29.85 | 23.48 | 28.68 | 21.85 | 32.87 | 25.78 | 27.08 |
| #1 | ✓ | ✗ | 30.45 | 24.11 | 29.41 | 22.53 | 33.41 | 25.85 | 27.63 |
| #2 | ✓ | ✓ | **30.68** | **24.12** | **29.65** | **22.78** | **33.65** | **26.18** | **27.84** |

Table 4: Ablation of Code Switch and word replacement strategies

- **Vecmap:** For the Vecmap method, we also test its performance on both supervised and unsupervised methods, as well as two distance metrics, Cosine and CSLS. Vecmap mainly includes the steps of Embedding normalization, whitening, orthogonal mapping, re-weighting, and self-learning. In the selection of the seed dictionary, the ground truth dictionaries published by the Muse are also used.

- **CSCBLI:** A method combing the contextual word representations and static word embeddings. Use the contextual word representations to adjust the static word embedding, then map the contextual word representation and calculate the similarity, calculate the similarity of the adjusted static word embedding, and then interpolate the two similarities. CSCBLI combing the contextual word representations and static word embeddings. Use the contextual word representations to adjust the static word embedding, then map the contextual word representations and calculate the similarity, calculate the similarity of the adjusted static word embeddings, and then interpolate the two similarities.

### 4.3 Setup

We evaluate the baseline systems and method we proposed on the language pair of En-Zh and Zh-En. All static word embeddings include the monolingual word embeddings and the mapped monolingual word embeddings in the experiment are 300-dimensional, and contextual word representations are 1024-dimensional. The Pre-trained model used in the experiment is the XLM model released by Facebook, in order to ensure the consistency of the data, we use the latest version of the wiki to fine-tune it. For the acquisition and processing of contextual word representations for each word, we randomly select ten sentences containing this word to calculate contextual word representations. If the total number of sentences containing this word is less than ten, we use all the sentences containing this word to do the calculation. And we select the best hyperparameters by searching a combina-tion of replacement ratio with the following range: token and sentence replacement ratio: {0.4; 0.5; 0.6; 0.8; 0.9; 1.0}, add word replacement strategies ratio:{0.1,0.2,0.3,0.4,0.5,0.6,0.7,0.8,0.9,1.0}.

|              | Medicine |       | Law   |       | Financial |       |
|--------------|----------|-------|-------|-------|-----------|-------|
|              | en-zh    | zh-en | en-zh | zh-en | en-zh     | zh-en |
| XLM_layer0   | 0        | 0     | 29.11 | 18.79 | 31.46     | 25.80 |
| XLM_layer1   | 30.68    | 24.12 | 29.65 | 22.78 | 33.65     | 26.18 |

Table 5: Performances of XLM_layer0 and XLM_layer1 to get the word representations used in proposed method

|                                | Medicine |        | Law    |        | Financial |        | avg   |
|--------------------------------|----------|--------|--------|--------|-----------|--------|-------|
|                                | En-zh    | Zh-en  | En-zh  | Zh-en  | En-zh     | Zh-en  |       |
| Muse_unsupervised$_{cosine}$   | 0        | 17.39  | 0      | 18.92  | 0         | 18.97  | 9.21  |
| Muse_supervised$_{cosine}$     | 14.63    | **19.65** | 9.73  | 14.50  | 15.27     | 18.85  | 15.44 |
| Muse_unsupervised$_{CSLS}$     | 0        | 17.95  | 0      | 18.92  | 0         | 19.74  | 9.44  |
| Muse_supervised$_{CSLS}$       | 19.64    | 16.52  | 11.73  | 14.50  | 18.19     | 16.15  | 16.12 |
| Vecmap_unsupervised$_{cosine}$ | 24.56    | 16.38  | 24.40  | 16.39  | 30.41     | 21.92  | **22.34** |
| Vecmap_supervised$_{cosine}$   | 19.95    | 17.82  | **25.73** | **19.04** | 19.39  | 17.69  | 19.94 |
| Vecmap_unsupervised$_{CSLS}$   | **27.54** | 18.82 | 13.33  | 14.63  | **30.54** | **23.59** | 21.41 |
| Vecmap_supervised$_{CSLS}$     | 22.93    | 13.51  | 14.00  | 11.10  | 19.92     | 13.08  | 15.76 |

Table 6: Performances of Muse and Vecmap with different distance metrics on domain datasets.

### 4.4 Main results

We report the main results of the BLI task on all test sets in table 3. As it can be seen that our method is based on the CSCBLI, so it also has a good performance, and the table1 shows that the method we proposed outperforms the strongest baseline CSCBLI at this task by an average of 0.76 percentage points.

For the three domain datasets, the Law dataset has the highest improvement, compared with CSCBLI, we have improved by 0.97 percentage points in the direction of en-zh, and 0.93 percent in the direction of zh-en, and for the other two datasets, there are also 0.74 and 0.59 percentage points of improvement, respectively. This shows that our method is effective on the BLI of the domain datasets, that is, it will be more suitable for the BLI based on the domain datasets.

## 5 Analysis

### 5.1 The influence of Code Switch and word replacement

In order to further explore the effects brought by the two components of the proposed method: Code Switch and word replacement strategy, we incrementally introduced it into the fine-tuning phases of the Pre-trained models in turn. table 4 presents the results of ablation experiments on three domain datasets. It can be seen that when only the Code Switch is used, there are only 0.55 scores promotions on average, and when the word replacement strategy is added to the Code Switch method, there are further 0.21 promotions on average. This shows that the proposed method has a more obvious effect on the BIL task than simply using the Code Switch method in the fine-tuning stage of the Pre-trained model.

### 5.2 Performances of the word representation without the XLM encoder layer and through the first layer of the encoder layer

Since CSCBLI only tested the performance of contextual word representation in different encoder layers, we naturally wondered whether the static word representation of the Pre-trained model can also bring a certain improvement effect. With such a conjecture, we conducted comparative experiments, using static word representation of XLM and dynamic first-level word representation of XLM to conduct experiments respectively.

As we can see in table 5, in our proposed method, the experimental accuracy used by the word representations of the layer$_0$ of XLM is lower than that of the layer₁ of it. However, except for the experimental result of medicine, the result on Law and Financial used the static word representations of XLM are not much worse than the contextual ones. This may give us a piece of information that using the word representations obtained by the Pre-trained model, even in the non-dynamic situation, can also adjust the static word embeddings gotten from other methods to perform better on the BLI task.

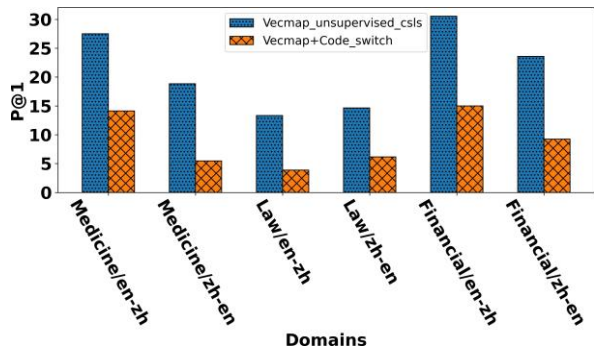

Figure 2: Performances of the Code_switch used in the static word embeddings training stage

### 5.3 Performances of the word representation without the XLM encoder layer and through the first layer of the encoder layer

In addition, to use the proposed method in dynamic word embeddings, we also tested adding the proposed method to the training process of static word embeddings. Corresponding to the fine-tuning process of the Pre-trained model, we tested the proposed method in the word embedding training of FASTTEXT and then trained the Vecmap after obtaining the new word embedding. The experimental results are shown in

## 6 Discussion

We added the Code Switch method to the Pre-trained model in the fine-tuning stage of the domain data-sets and added a word replacement strategy on this basis. Our analysis shows that this method can make the word embedding more suitable for the BLI method, that is, in a unified space, the two languages are better aligned, resulting in higher quality translated word pairs. Under the conditions of the domain corpus, the word frequency in the corpus is lower, and the performance of static word embedding is correspondingly reduced. However, the method of word replacement is used to pull the distance between the same words in different languages to a certain extent.

In addition, as shown in table 6, we note that for the Muse and Vecmap, their performance gap is relatively large compared to the representation in the general domain, and in some cases, the supervised method is less effective than the unsupervised method. For example, using the CSLS distance metric method under the medicine dataset, the Vecmap supervised method is 4.61 percentage points worse than the unsupervised method in the en-zh direction, and 5.31 percentage points in the zh-en direction. In the Muse method, the supervised method can break the 0 dilemmas in the en-zh language pair, but there are still situations where the supervised method is much worse than the unsupervised method in the zh-en direction, such as the Cosine measurement method on the law dataset. The unsupervised method is 4.42 percentage points lower than the supervised method, and the CSLS metric method is 1.43 percentage points lower on the medicine dataset. However, since this is not the focus of this work, we did not investigate this phenomenon in depth. This shows that supervised bilingual dictionary extraction will have a negative impact on the accuracy of vocabulary translation when the seed dictionary is given words that belong to a non-domain or general domain. At the same time, it also shows that there is still room for improvement in terms of domain vocabulary extraction, whether it is the method design of supervised bilingual dictionary extraction or the method of lexical similarity measurement, which also provides ideas and entry points for the improvement of BLI in the future.

## 7 Conclusion

Most BLI systems focus on general domain improvements, but no work is dedicated to cross domains. In this paper, we propose a mechanism based on the Code Switch and Pre-trained model, which replaces the vocabulary of the target language with the corresponding vocabulary of the source language in the fine-tuning stage of the Pre-trained model. In the method of Code Switch, the strategy of word replacement is added, and strategies to adapt to different contexts are formulated by judging whether the words belong to general words or specialized words. Experiments show that this method can pull in the distance between the corresponding word pairs between the two language pairs. Under the setting of the professional domain corpus, it achieves a certain improvement in the accuracy of word translation in the BLI task.